\documentclass[11pt]{article}

\usepackage{times}
\usepackage{url}
\usepackage{amsmath}
\usepackage{amsfonts}
\usepackage{hhline}
\usepackage{latexsym}
\usepackage{multirow}
\usepackage{graphics}
\usepackage{diagbox}

\usepackage{arxiv}
\usepackage{amsthm}
\usepackage{amssymb}
\usepackage{mathtools}
\usepackage{graphicx}
\usepackage{xcolor}
\usepackage{subfigure}
\usepackage{hyperref}

\newcommand{\bigcdot}{\boldsymbol{\cdot}}

\title{\bf {\LARGE NePTuNe:} \\ Neural Powered Tucker Network \\ for Knowledge Graph Completion}
\author{Shashank Sonkar \\
  Rice University \\
  {\tt ss164@rice.edu}
  \And
  Arzoo Katiyar \\
  Pennsylvania State University \\
  {\tt arzoo@psu.edu} 
  \And
  Richard G. Baraniuk \\
  Rice University \\
  {\tt richb@rice.edu}
}
  
\date{}

\begin{document}

\maketitle
\begin{abstract}
    Knowledge graphs link entities through relations to provide a structured representation of real world facts. However, they are often incomplete, because they are based on only a small fraction of all plausible facts. The task of {\em knowledge graph completion} via {\em link prediction} aims to overcome this challenge by inferring missing facts represented as links between entities. Current approaches to link prediction leverage tensor factorization and/or deep learning. 
    Factorization methods train and deploy rapidly thanks to their small number of parameters but have limited expressiveness due to their underlying linear methodology.
    Deep learning methods are more expressive but also computationally expensive and prone to overfitting due to their large number of trainable parameters. 
    We propose \textit{Neural Powered Tucker Network} (NePTuNe), a new hybrid link prediction model that couples the expressiveness of deep models with the speed and size of linear models. 
    We demonstrate that NePTuNe provides state-of-the-art performance on the FB15K-237 dataset and near state-of-the-art performance on the WN18RR dataset.
\end{abstract}
\vspace{0.1cm}

\section{Introduction}
Knowledge graphs encode information by means of triples of the form $(h, r, t)$, where relation $r$ represents the link between the head entity $h$ and tail entity $t$. This rich structured information is useful for a wide range of natural language processing tasks, e.g., question answering, multi-hop reasoning, and information retrieval \cite{hao2017end,zhang2018variational,dietz2018utilizing}. However, knowledge graphs are often incomplete, meaning that they contain only a small subset of all known connections. Thus the task of {\em knowledge graph completion} via {\em link prediction} is of great importance, since populating knowledge graphs manually is both arduous and costly.

There are currently two families of link prediction methods at opposite ends of the expressiveness vs.\ scale/speed tradeoff.
Tensor factorization methods represent the knowledge graph as a 3-mode binary tensor \cite{nickel2011three,yang2014embedding,trouillon2017knowledge,balavzevic2019tucker}. 
These methods are easy and fast to train and deploy thanks to their small number of parameters.
However, they have limited expressiveness due to their underlying linear methodology.  
Deep learning methods are more expressive but are computationally expensive and prone to overfitting due to their large number of trainable parameters  \cite{nickel2011three,socher2013reasoning}. 

The state-of-the-art linear model for link prediction is TuckER \cite{balavzevic2019tucker}, which is based on the Tucker decomposition \cite{tucker1966some,tucker1964extension} of a third-order binary tensor. 
Popular linear models like RESCAL \cite{nickel2011three}, DistMult \cite{yang2014embedding}, ComplEx \cite{trouillon2016complex}, and SimplE \cite{poole2018simple} are special cases of TuckER. 
The pioneering deep learning model for link prediction is the Neural Tensor Network (NTN) \cite{socher2013reasoning} which offers high expressiveness but comes with a major shortcoming.
Since it uses a different weight matrix for modeling each relation, its number of parameters explodes with the graph size, resulting in computationally expensive training and overfitting issues.


In this paper, we propose \textit{Neural Powered Tucker Network} (NePTuNe), a hybrid link prediction model that leverages desirable features from both TuckER and NTN. 
Carefully crafted nonlinearities make NePTuNe considerably more expressive than TuckER. 
Harnessing the principles of a shared core tensor intrinsic to the Tucker decomposition leads to parameter-sharing which simultaneously significantly reduces the number of parameters, remedies overfitting, and makes NePTuNe fast and efficient to train.
Even more importantly, the Tucker formulation enables NePTuNe to be trained using the {\em 1-$N$ loss function} by employing the re-arrangement property of $n$-mode multiplication. 
NePTuNe's capacity to use 1-$N$ loss function is crucial, since the loss function has been shown to be greatly advantageous in training link prediction models.
Importantly, deep learning models like NTN are incompatible with the 1-$N$ loss.

Our experimental evaluations demonstrate that NePTuNe outperforms state-of-the-art linear and deep learning models on the FB15k-237 dataset \cite{toutanova2015representing} and achieves second-best performance on the WN18RR \cite{dettmers2018convolutional} dataset. 
We also show that NePTuNe has the same space and training and testing time complexity as the state-of-the-art linear TuckER model. 
Hence, NePTuNe successfully balances the expressiveness vs.\ scale/speed tradeoff introduced above.
Code and instructions to reproduce our results are available \textcolor{blue}{\href{https://github.com/luffycodes/neptune}{here}}.

\section{Background}

Following \cite{balavzevic2019tucker}'s notation, let $\mathcal{E}$ be the set of entities, $\mathcal{R}$ the set of relations, and $\mathcal{K} \subseteq \mathbb{K} = \mathcal{E} \times \mathcal{R} \times \mathcal{E}$ the knowledge graph. $(h, r, t) \in \mathcal{K}$ denotes a triple in the knowledge graph, where $h, t \in \mathcal{E}$ and $r \in \mathcal{R}$. Let the scoring function $\phi(h, r, t) : \mathcal{E} \times \mathcal{R} \times \mathcal{E} \rightarrow \mathbb{R}$ measure the plausibility of a triple $(h, r, t) \in \mathcal{K}$. Boldface lower-case letters $\boldsymbol{\mathrm{x}}$ denote a vector, while boldface upper-case letters $\boldsymbol{\mathrm{X}}$ denote a matrix. Script upper-case letters $\mathrm{\mathcal{X}}$ denote a three-mode tensor. The embeddings for the entities and relations are given by the matrices $\mathbf{E} \in \mathbb{R}^{|\mathcal{E}| \times d}$ and $\mathbf{W} \in \mathbb{R}^{|\mathcal{R}| \times k}$, respectively.

We now introduce the TuckER and NTN models. 
As mentioned above, TuckER is based on the Tucker decomposition of a 3-mode tensor $\mathcal{X}$, which is given by
\begin{equation}
    \mathcal{X} \approx \mathcal{G} \times_1 \mathbf{A} \times_2 \mathbf{B} \times_3 \mathbf{C},
    \label{eq:decompose}
\end{equation}
where $\mathcal{X} \in \mathbb{R}^{I \times J \times K}, \mathcal{G} \in \mathbb{R}^{L \times M \times N}, \mathbf{A} \in \mathbb{R}^{I \times L}, \mathbf{B} \in \mathbb{R}^{J \times M}, \mathbf{C} \in \mathbb{R}^{K \times N}$, and $\times_n$ is the $n$-mode product. 
We can write TuckER's scoring function as
\begin{equation}
    \phi_{\text{TuckER}}(h, r, t) = \mathrm{\mathcal{W}} \times_1 \mathbf{e}_h \times_2 \mathbf{w}_r \times_3 \mathbf{e}_t,
    \label{eq:tucker_eq}
\end{equation}
where $\mathbf{w}_r$, $\mathbf{e_h, e_t} \in \mathbb{R}^d$ are embeddings for the relation, head, and tail entities, and $\mathrm{\mathcal{W}} \in \mathbb{R}^{d \times k \times d}$ is the shared core tensor. Table~\ref{tab:results_space} lists the scoring functions for some additional linear link prediction models.

NTN's scoring function is given by
\begin{equation}
    \phi_{\text{NTN}}(h, r, t) = \boldsymbol{\mathrm{w}}_r^T \bigcdot f_{\rm act}(\mathrm{\mathcal{W}}_r \times_1 \mathbf{e}_h \times_3 \mathbf{e}_t),
    \label{eq:ntn_eq}
\end{equation}
where $\mathcal{W}_r \in \mathbb{R}^{d \times d \times k}$ is the relation specific tensor, $\bigcdot$ denotes the dot (inner) product, and $f_{\rm act}$ is a nonlinear scalar activation function such as ReLU or $\tanh$.

Observe that NTN can be viewed as a nonlinear version of TuckER (albeit with a different core tensor per relation). 
By comparing (\ref{eq:ntn_eq}) with (\ref{eq:tucker_as_ntn}), TuckER's scoring function can be written as NTN's scoring function but without a nonlinearity as
\begin{align}
    \phi_{\text{tuckER}}(h, r, t) &= \mathrm{\mathcal{W}} \times_1 \mathbf{e}_h \times_2 \mathbf{w}_r \times_3 \mathbf{e}_t \label{eq:tucker_as_ntn1}\\
    &= \mathrm{\mathcal{W}} \times_1 \mathbf{e}_h \times_3 \mathbf{e}_t \times_2 \mathbf{w}_r \label{eq:tucker_as_ntn2}\\
    &= \mathbf{w}_r^T \bigcdot (\mathrm{\mathcal{W}} \times_1 \mathbf{e}_h \times_3 \mathbf{e}_t) \label{eq:tucker_as_ntn},
\end{align}
where (\ref{eq:tucker_as_ntn2}) follows from the re-arrangement property of $n$-mode multiplication, and (\ref{eq:tucker_as_ntn}) follows from the properties of the dot product and $n$-mode multiplication. 
Viewing NTN against this backdrop of the Tucker decomposition sets the stage for defining a single shared core tensor $\mathcal{W}$ across relations, which is one of key defining traits of our proposed model, NePTuNe.

\section{NePTuNe: Neural Powered Tucker Network}

\subsection{The NePTuNe Model}

The scoring function of our proposed model NePTuNe is given by
\begin{equation}
    \phi_{\text{NePTuNe}}(h, r, t) = \mathbf{e}_t^T \bigcdot f_{\rm act}(\mathrm{\mathcal{W}} \times_1 \mathbf{e}_h \times_2 \mathbf{w}_r).
    \label{eq:NePTuNe_eq}
\end{equation}
%
Like NTN, NePTuNe utilizes a nonlinear activation function $f_{\rm act}$.
However, unlike NTN (which employs a different $\mathcal{W}_r$ for each relation $r \in \mathcal{R}$) and inspired by TuckER, NePTuNe uses a shared $\mathcal{W}$ across all relations.  
The shared core tensor $\mathcal{W}$ enables parameter-sharing among relations, which mitigates the problem of overfitting that plagues NTN and accelerates training.

Rewriting the TuckER scoring function from \eqref{eq:tucker_as_ntn1} as
\begin{align}
    \phi_{\text{tuckER}}(h, r, t) 
    &= \mathbf{e}_t^T \bigcdot (\mathrm{\mathcal{W}} \times_1 \mathbf{e}_h \times_2 \mathbf{w}_r) \label{eq:tucker_as_neptune}
\end{align}
and comparing to \eqref{eq:NePTuNe_eq}, we see that the NePTuNe scoring function shares the exact same structure save for the application of the nonlinear $f_{\rm act}$ to the second term.
NePTuNe also shares the same structure as NTN except that the order of $\mathbf{e}_t$ and $\mathbf{w}_r$ is reversed.
This turns out to be critically important, because it enables us to apply the {\em $1-N$ loss function} to train NePTuNe.
We discuss this at length below in Section \ref{complexity}. 
In the absence of the nonlinearity ($f_{\rm act}$ set to identity), the NePTuNe \eqref{eq:NePTuNe_eq} and NTN \eqref{eq:ntn_eq} scoring functions are equivalent, since the $n$-th mode product in the Tucker decomposition is invariant to the order [($\mathrm{\mathcal{W}} \times_1 \mathbf{e}_h \times_2 \mathbf{w}_r) \times_3 \mathbf{e}_t = (\mathrm{\mathcal{W}} \times_1 \mathbf{e}_h \times_3 \mathbf{e}_t) \times_2 \mathbf{w}_r$]. 
However, with the nonlinearity, the scoring functions are not equivalent [$f_{\rm act}(\mathrm{\mathcal{W}} \times_1 \mathbf{e}_h \times_2 \mathbf{w}_r) \times_3 \mathbf{e}_t \neq f_{\rm act}(\mathrm{\mathcal{W}} \times_1 \mathbf{e}_h \times_3 \mathbf{e}_t) \times_2 \mathbf{w}_r$].


\begin{table*}[t]
\centering
\resizebox{\columnwidth}{!}{
\begin{tabular}{llll}
\hline
Model &  Scoring Function   & Relation Parameters & Space Complexity  \\ \hline
RESCAL \cite{nickel2011three}  &  $\mathbf{e}_h^T\mathbf{W}_r\mathbf{e}_t$   & $\mathbf{W}_r \in \mathbb{R}^{d^2}$ & $\mathcal{O}(\mathcal{|E|} d + \mathcal{|R|} k^2)$  \\

DistMult \cite{yang2014embedding}  &  $\langle \mathbf{e}_h,\mathbf{w}_r,\mathbf{e}_t \rangle$   & $\mathbf{w}_r \in \mathbb{R}^{d}$ & $\mathcal{O}(\mathcal{|E|} d + \mathcal{|R|} d)$  \\

ComplEX \cite{trouillon2016complex}  &  $\mathrm{Re}(\langle \mathbf{e}_h,\mathbf{w}_r,\Bar{\mathbf{e}}_t \rangle)$   & $\mathbf{w}_r \in \mathbb{C}^{d}$ & $\mathcal{O}(\mathcal{|E|} d + \mathcal{|R|} d)$  \\

SimplE \cite{poole2018simple} & $\frac{1}{2}(\langle \mathbf{e}_h,\mathbf{w}_r,\mathbf{e}_t \rangle + \langle \mathbf{e}_h',\mathbf{w}_r,\mathbf{e}_t' \rangle)$ & $\mathbf{w}_r \in \mathbb{R}^{k}$ & $\mathcal{O}(\mathcal{|E|} d + \mathcal{|R|} d)$  \\

TuckER \cite{balavzevic2019tucker} & $\mathcal{W} \times_1 \mathbf{e}_h \times_2 \mathbf{w}_r \times_3 \mathbf{e}_t$ & $\mathbf{w}_r \in \mathbb{R}^{k}$ & $\mathcal{O}(\mathcal{|E|} d + \mathcal{|R|} k)$ \\

NePTuNe (ours) & $\mathbf{e}_t^T \bigcdot f_{\rm act}(\mathrm{\mathcal{W}} \times_1 \mathbf{e}_h \times_2 \mathbf{w}_r)$ & $\mathbf{w}_r \in \mathbb{R}^{k}$ & $\mathcal{O}(\mathcal{|E|} d + \mathcal{|R|} k)$ \\

\hline
\end{tabular}
}
\caption{Scoring function and space complexity (significant terms) of state-of-the-art linear link prediction models in comparison to NePTuNe. 
Information gathered from \protect\cite{balavzevic2019tucker}.}
\label{tab:results_space}
\end{table*}

\subsection{Training}

We train NePTuNe using a binary cross entropy (BCE) loss with $1-N$ scoring.
$1-N$ scoring was proposed by \cite{dettmers2018convolutional} to speed up training and performance of link prediction models. For an arbitrary head entity $h$ and relation $r$, the $1-N$ score with BCE loss is given by
\begin{equation}
    L_{(h,r)} = - \sum_{i=1}^{|\mathcal{E}|}\left[ y_i \mathrm{log}(p_i) + (1-y_i)\mathrm{log}(1-p_i) \right],
    \label{eq:loss_NePTuNe}
\end{equation}
where $y_i=1$ for all tail entities $t_i \in \mathcal{E}$ for which the triple $(h, r, t_i) \in \mathcal{K}$ and $y_i=0$ otherwise, and $p_i = \mathrm{sigmoid}(\phi_{\mathrm{NePTuNe}}(h, r, t_i))$. 
We also apply the inverse relation data-augmentation technique \cite{dettmers2018convolutional,lacroix2018canonical}, which adds a new triple $(t, r^{-1}, h)$ in $\mathcal{K}$ for every $(h, r, t) \in \mathcal{K}$ by defining a new relation $r^{-1}$ for each relation $r$.

\subsection{Time Complexity Analysis}
\label{complexity}

We study the number of multiplication operations needed to compute $L_{(h,r)}$ in \eqref{eq:loss_NePTuNe} for an arbitrary head entity $h$ and relation $r$.  
The analysis can be reduced to computing the number of multiplication operations needed to compute the scoring function $\phi_{\rm NePTuNe}(h,r,t)$ for all $t \in \mathcal{E}$.

Define the scoring function vector 
$\mathbf{\mathrm{\Phi}}_{\text{NePTuNe}}(h, r) \in \mathbb{R}^{|\mathcal{E}|}$ with $i^{th}$ element $\mathrm{\Phi}_{\text{NePTuNe}}^i(h, r) = \phi_{\mathrm{NePTuNe}}(h, r, t_i)$. Then,
\begin{align}
    \mathbf{\mathrm{\Phi}}_{\text{NePTuNe}}(h, r) =\ f_{\rm act}(\mathrm{\mathcal{W}} \times_1 \mathbf{e}_h \times_2 \mathbf{w}_r) \times_3 \mathbf{E}
     \\
     =\ \underbrace{f_{\rm act}(\underbrace{(\mathrm{\mathcal{W}} \times_2 \mathbf{w}_r)}_{\mathbf{A}} \times_1 \mathbf{e}_h )}_{\mathbf{b}} \times_3 \mathbf{E}
\end{align}
since $n$-mode multiplication is invariant to order. 
The operation $(\mathrm{\mathcal{W}} \times_2 \mathbf{w}_r)$ reduces the 3-D tensor $\mathcal{W} \in \mathbb{R}^{k \times d \times d}$ to a 2-D matrix $\mathbf{A} \in \mathbb{R}^{d \times d}$, which, followed by $\times_1 \mathbf{e}_h$ and the nonlinearity reduces $\mathbf{A}$ to a vector $\mathbf{b} \in \mathbb{R}^d$.

Operation $\mathrm{\mathcal{W}} \times_2 \mathbf{w}_r$ requires $\mathcal{O}(d \times d \times k)$ multiplications; operation $\mathbf{A} \times_1 \mathbf{e}_h$ requires $\mathcal{O}(d \times d)$ multiplications; and operation $\mathbf{b} \times_3 \mathbf{E}$ requires $\mathcal{O}(d \times \mathcal{|E|})$. 
Thus, computing $\mathbf{\mathrm{\Phi}}_{\text{NePTuNe}}(h, r)$ involves $\mathcal{O}(d \times d \times k + d \times d \times 1 + 1 \times d \times \mathcal{|E|})$ multiplications.

This analysis enables us to compare the effect of interchanging $\mathbf{w_r}$ and $\mathbf{e_t}$ to contrast the NePTuNe scoring function \eqref{eq:NePTuNe_eq} with the NTN scoring function \eqref{eq:ntn_eq}. 
The scoring function of NTN is given by
\begin{equation}
    \mathbf{\mathrm{\Phi}}_{\text{NTN}}(h, r) = \underbrace{f_{\rm act}(\underbrace{(\mathrm{\mathcal{W}}_r \times_1 \mathbf{e}_h)}_{\mathbf{C}} \times_3 \mathbf{E} )}_{\mathbf{d}} \times_2 \mathbf{w}_r.
\end{equation}
Computing $\mathbf{\mathrm{\Phi}}_{\text{NTN}}(h, r)$ requires $\mathcal{O}(d \times d \times k + d \times d \times \mathcal{|E|} + k)$ multiplications. 
The bottleneck operation $\mathbf{C} \times_3 \mathbf{E}$ that requires $d \times d \times \mathcal{|E|}$ multiplications is what makes $1-N$ scoring infeasible for NTN.

The above time complexity analysis reveals that the design decision to interchange the order of the terms in the NePTuNe scoring function \eqref{eq:NePTuNe_eq} vs.\ the NTN \eqref{eq:ntn_eq} is significant, since it enables NePTuNe to use $1-N$ scoring, which has proven effective in accelerating the training of link prediction models.

\begin{table*}[t]
\centering
\resizebox{\columnwidth}{!}{
\begin{tabular}{lllll|llll}
 & \multicolumn{4}{c}{FB15K-237} & \multicolumn{4}{c}{WN18RR}    \\ 
\cline{2-5} \cline{6-9} 
&  MRR   & Hits@1 & Hits@3 & Hits@10 &  MRR   & Hits@1  & Hits@3 & Hits@10 \\    \hline
DistMult  &  .241   & .155 & .263 & .419 &  .430   & .390 & .440 & .490 \\
ConvE  &  .325   & .237 & .356 & .501 &  .430   & .400 & .440 & .520 \\
TuckER  &  .358   & .266 & .394 & .544 &  .470   & .443 & .482 & .526 \\
MurE  &  .336   & .245 & .370 & .521 &  .475   & .436 & .487 & .554 \\
ComplEx-N3  &  .357   & .264 & .392 & .547 &  .480   & .435 & .495 & .572 \\
RotatE  &  .338   & .241 & .375 & .533 &  .476   & .428 & .492 & .571 \\
Quaternion  &  .348   & .248 & .382 & \textbf{.550} &  .488   & .438 & .508 & .582 \\
MurP  &  .335   & .243 & .367 & .518 &  .481   & .440 & .495 & .566 \\
RotH  &  .344   & .246 & .380 & .535 &  \textbf{.496}   & .449 & \textbf{.514} & \textbf{.586} \\
NePTuNe (ours) & \textbf{.366} & \textbf{.272} & \textbf{.404} & .547 & .491 & \textbf{.455} & .507 & .557 \\
\end{tabular}
}

\caption{Evaluation results comparing NePTuNe with a range of state-of-the-art linear and nonlinear link prediction models. 
Results for all methods other than NePTuNe are taken from the RotH paper \protect\cite{chami2020low}. 
NePTuNe outperforms the state-of-the-art linear TuckER model on the FB15k-237 dataset by a significant margin while also outperforming the state-of-the-art nonlinear RotH model on the WN18RR dataset in the Hits@1 metric. 
Importantly, NePTuNe makes no assumptions about the data, and hence its performance is consistent across both datasets. 
RotH, in contrast, was specifically designed for hierarchical datasets like WN18RR, which explains its subpar performance on FB15k-237.
}
\label{tab:results_ntn2}
\end{table*}

\subsection{Space Complexity Analysis}
\label{space_complexity}

The parameters of NePTuNe are the shared core tensor $\mathcal{W} \in \mathbb{R}^{d \times d \times k},$ entity embeddings $\mathbf{E} \in \mathbb{R}^{\mathcal{|E|} \times d}$, and relation embeddings $\mathbf{W} \in \mathbb{R}^{\mathcal{|R|} \times k}$, leading to the significant terms of NePTuNe's space complexity being of $\mathcal{O}(|\mathcal{E}| \times d + |\mathcal{R}| \times k)$. 

Since the above space and time complexity analysis also holds for TuckER, we have verified that NePTuNe and TuckER are equivalent in these respects.

\section{Experiments}

We compare the performance of NePTuNe against a wide range of state-of-the-art linear and nonlinear link prediction models, including ConvE\cite{dettmers2018convolutional}, MurP \cite{balazevic2019multi}, 
MurE \cite{balazevic2019multi}, 
RotatE \cite{sun2019rotate}, 
ComplEx-N3 \cite{lacroix2018canonical}, 
Quaternion \cite{NEURIPS2019_d961e9f2},
TuckER \cite{balavzevic2019tucker}, 
and RotH \cite{chami2020low} on the standard datasets 
FB15k-237 \cite{toutanova2015representing} and WN18RR \cite{dettmers2018convolutional}. 
FB15k-237 has 14,541 entities and 237 relations, while WN18RR has 40,943 entities and 11 relations.


We extended the TuckER codebase \cite{tucker_code} and make the code publicly available with the hyper-parameters needed to replicate the experiments \textcolor{blue}{\href{https://github.com/luffycodes/neptune}{here}}.
As with TuckER, we use batch normalization \cite{ioffe2015batch} and dropout \cite{srivastava2014dropout} on the input embeddings and hidden representations for regularization. 
We optimize NePTuNe in terms of the  the $1-N$ score with BCE loss from \eqref{eq:loss_NePTuNe}
using Adam \cite{kingman2015adam} for 1000 training epochs. 
(We found that ReLU performed marginally better than $\tanh$ for the activation $f_{\rm act}$.)
We measure performance using the standard \textit{filtered} metrics of mean reciprocal rank (MRR), Hits@1, Hits@3, and Hits@10 \cite{bordes2013translating}. MRR is defined as the mean over all test triples of the inverse of the rank assigned to the true triple tensor. Hits@$n$ is defined as the fraction of times the true triple is present among the top $n$ ranked candidate triples.



Table \ref{tab:results_ntn2} compares NePTuNe with a range of state-of-the-art linear and nonlinear link prediction models.
We see that NePTuNe has superior performance on FB15k-237 and outperforms the state-of-the-art nonlinear RotH model \cite{chami2020low} in the Hits@1 metric on WN18RR.
In contrast to RotH, which has been specifically designed to model the kinds of hierarchical relations that feature prominently in WN18RR, NePTuNe's model design is dataset-agnostic. 
This explains its consistent performance on both FB15k-237 and WN18RR.
%
%
Finally, it is worth noting that NePTuNe has the same number of parameters as the state-of-the-art linear TuckER model yet significantly outperforms it on both datasets due to its increased expressive power.

\section{Conclusions} 

We have developed NePTuNe, a compact and computationally efficient model for knowledge graph completion that combines the speed and robustness of linear with the expressivity of nonlinear approaches to link prediction.
An important hallmark of NePTuNe is that it is dataset-agnostic, as demonstrated by its  (near) state-of-the-art performance on two very different standard datasets.



\section*{Acknowledgements}

This work was supported by NSF grants CCF-1911094, IIS-1838177, IIS-1730574, and IUSE-1842378; ONR grants N00014-18-12571, N00014-20-1-2787, and N00014-20-1-2534; AFOSR grant FA9550-18-1-0478; and a Vannevar Bush Faculty Fellowship, ONR grant N00014-18-1-2047.


\bibliographystyle{acl}
\bibliography{main}

\end{document}